# Nuclear Norm based Matrix Regression with Applications to Face Recognition with Occlusion and Illumination Changes

Jian Yang*, Jianjun Qian, Lei Luo, Fanlong Zhang, Yicheng Gao

School of Computer Science and Engineering, Nanjing University of Science and Technology, Nanjing 210094, P. R. China

*Corresponding author

E-mail: csjyang@njust.edu.cn (Jian Yang)

**Abstract**: Recently regression analysis becomes a popular tool for face recognition. The existing regression methods all use the one-dimensional pixel-based error model, which characterizes the representation error pixel by pixel individually and thus neglects the whole structure of the error image. We observe that occlusion and illumination changes generally lead to a low-rank error image. To make use of this low-rank structural information, this paper presents a two-dimensional image matrix based error model, i.e. matrix regression, for face representation and classification. Our model uses the minimal nuclear norm of representation error image as a criterion, and the alternating direction method of multipliers method to calculate the regression coefficients. Compared with the current regression methods, the proposed Nuclear Norm based Matrix Regression (NMR) model is more robust for alleviating the effect of illumination, and more intuitive and powerful for removing the structural noise caused by occlusion. We experiment using four popular face image databases, the Extended Yale B database, the AR database, the Multi-PIE and the FRGC database. Experimental results demonstrate the performance advantage of NMR over the state-of-the-art regression based face recognition methods.

# 1. Introduction

Face recognition has aroused broad interests in pattern recognition and computer vision areas in the past 20 years. Meanwhile numerous face representation and classification methods have been developed. Recently, linear regression (LR) analysis based methods have become a hot topic in face recognition community. I. Naseem et al. presented a linear regression classifier (LRC) for face classification [1]. Actually, the previous works, like the nearest feature line [2], the nearest feature plane and the nearest feature space methods [3] are all variants of LR based methods.

To avoid over-fitting, a regularization term is generally imposed to the LR model. There are two widely-used regularizers: the $L_2$-norm based regularizer and the $L_1$-norm based one. LR with the $L_2$-norm regularizer is generally called Ridge regression, while LR with the $L_1$-norm regularizer is called Lasso, which is a popular model for sparse representation. J. Wright et al. [4] presented a sparse representation based classification (SRC) method. SRC uses all training samples as the dictionary to represent a test sample and assumes the representation coefficients are sparse; the sparse nonzero representation coefficients are supposed to concentrate on the training samples with the same class label as the test sample. To obtain a robust model, they further assume noises are sparse and give the extended SRC model. The model shows the strong ability of dealing with sparse random pixel corruption and block occlusion. A. Wagner et al. [10] further extended the SRC model and unified the face alignment and recognition into one framework.

Some recent work, on the other hand, began to investigate the role of sparsity in face recognition [13-16]. J. Yang et al. [15] gave an insight into SRC and provided some theoretical supports for its effectiveness. They argued that it is $L_1$ constraint rather than $L_0$ (the inherent sparse constraint) that makes SRC effective. L. Zhang et al. [16] analyzed the working principle of SRC and believed that the collaborative representation strategy plays a more important role than the $L_1$-norm based sparsity constraint. They presented a collaborative representation classifier (CRC) based on Ridge regression. CRC, however, does not provide a mechanism for noise removing, so it is not a robust method for face recognition.

In the LRC, CRC and SRC models, the representation residual is measured by the $L_2$-norm or $L_1$-norm of the error vector. Such models inherently assume that the representation error follows Gaussian or Laplacian distribution. However, in real world face recognition cases, the distribution of representation error is more complicated [6, 7]. So, in theory, the models mentioned above are not sufficiently robust for practical noises. Towards this end, M. Yang et al. borrowed the idea robust regression [5] and proposed a regularized robust coding method [6, 7]. Based on the correntropy induced robust error metric, R. He et al. presented the correntropy based sparse representation (CESR) algorithm [8, 9]. It is interesting that although CESR and RSC are developed from different motivations, they are both in spirit of a sparse robust regression model, noticing that correntropy can be viewed as an M-estimator with varying kernel sizes. Recently, in order to unify the two kinds of existing sparse robust regression models: the additive model represented by SRC for error correction and multiplicative model represented by CESR and RSC for error detection, R. He et al. [34] built a half-quadratic framework: by defining different half-quadratic functions, the framework enables to perform both error correction and error detection. In addition, Naseem et al. further extended their LRC to the robust linear regression classification (RLRC) using the Huber estimator to deal with severe random pixel noise and illumination changes [17]. All of these robust regression related methods have been applied to real-world face recognition problems and yielded promising results.

The existing robust regression methods all use the one-dimensional pixel-based error model. That is, the error on each pixel is characterized one by one individually. Specifically, given a dictionary $D = [d_1; d_2; …; d_n]$, where $d_i$ is the $i$th row of $D$. Let $e = y - D\alpha = [e_1; e_2; …; e_m]$, where $e_i = y_i - d_i\alpha$, $i=1,2,…,m,$ and $m$ is the number of pixels. The one-dimensional pixel-based error model has two problems. First, this modeling assume that $e_1, e_2, …, e_m$ are independent and identically distributed (i.i.d.). This assumption is no problem for the random pixel corruption, where the noise is added independently on each pixel. However, in many cases of occlusion, this assumption does not hold. For instance, in the black scarf caused occlusion part, pixel values are zeros. So, the ideal representation errors in the occluded part are correlated, because pixels in a local area of a real-world image are generally highly-correlated. A similar example is shown in Fig.1, where for a given face image, we generate 100 face images each with a randomly located 45x45 occlusion part. The correlation map of pixels in the recovered occluded part is shown in Fig. 1, from which we can see that these pixels are obviously not independent. Actually, for more complicated occlusion like that cased by a real object, we can get similar results. Therefore, using the one-dimensional pixel-based error model (such as SRC [4], RSC [6, 7], Robust LRC [17] etc.) to address image classification with occlusions is theoretically questionable.

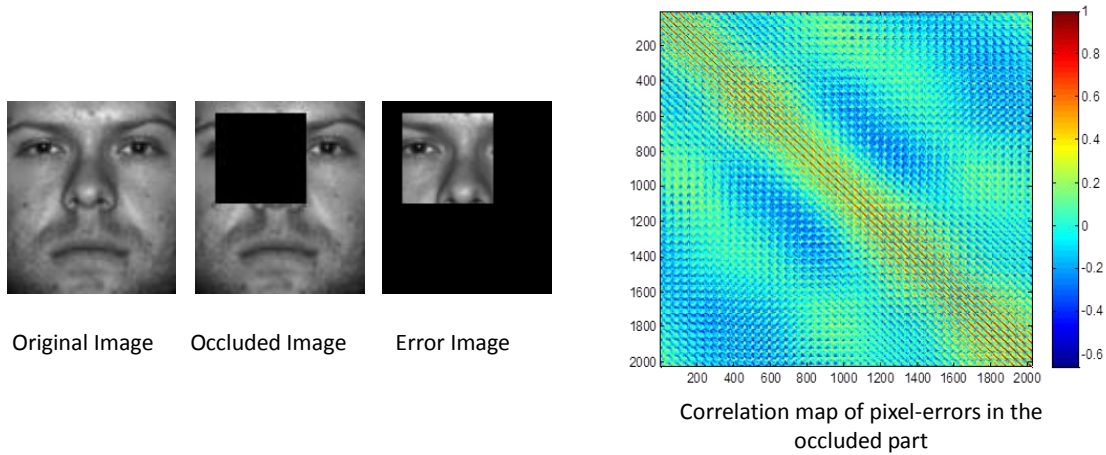

Original Image    Occluded Image    Error Image

Correlation map of pixel-errors in the occluded part

**Fig.1.** An example that shows the pixel-errors in occluded part are correlated

Second, characterizing the representation error pixel by pixel individually neglects the whole structure of the error image, noting that all pixel-errors form an error image which may contain meaningful structural information (e.g. the rank of error image). In regression analysis based face recognition methods, we use training images to represent a test image. Ideally, the error image is a zero matrix. In this case, the error image is naturally low-rank. In more general cases, there exist illumination variations and occlusions in test image. Actually, illumination and occlusion are two critical factors that affect the performance of face recognition. In practice, illumination changes, particularly partial illumination variations such as shadows, generally lead to a low-rank error image, in contrast to the full-rank original image. Occlusion, such as sunglass and scarf, also yields a low-rank error image. Some examples of error images (residual images) yielded by illumination changes and occlusion are shown in Figs. 1, 3 and 4. The existing regression methods, characterizing the pixel-error individually, fail to utilize the kind of structural information.

To make full use of the low-rank structural information of error image, this paper presents a *two-dimensional image matrix* based error model, i.e. matrix regression, to carry out the image representation and classification straightforward. In contrast, the previous methods, ridge regression, Lasso or robust regression, are all vector-based regression model. That is, for dealing with 2D image in the form of matrices, we have to convert images into vectors in advance before using this kind of regression models. In the converting step, some structural information (e.g. the rank of error image) might be lost. Our matrix regression model does not need the matrix-to-vector converting step. It uses the structural information of images by minimizing the rank of representation residual image to determine the regression coefficients.

The rank minimization problem, however, is difficult to solve since it is NP-hard in general. Fazel et al. [18, 19] found that the convex envelope of rank (X) is the nuclear norm of X, i.e. $\|\mathbf{X}\|_*$, which is the sum of singular values of X. They gave the nuclear norm heuristic for solving the rank minimization problem. Recent research efforts provided theoretical analysis and conditions for exact recovery of low-rank matrices via nuclear norm optimization [20-23]. Based on these results, Candès [24] and Wright et al. [25] proposed the Robust PCA (RPCA) method. G. Liu [39] et al. proposed the low-rank representation (LRR) for subspace segmentation. Both methods assume the noise is sparse, but RPCA assumes the clean data matrix is low rank while LRR assumes the representation coefficient matrix is low rank. In addition, R. He et al. presented a method for recovering the corrupted low-rank matrices via half-quadratic based nonconvex minimization [26]. Z. Liu [27] et al. proposed the interior-point method for nuclear norm approximation with application to system identification. In a similar spirit, in this paper, we use the minimal nuclear norm of the representation residual image as a criterion in our matrix regression model. So, our method is named nuclear norm based matrix regression (NMR).

Compared with the state-of-the-art regression based classification methods, the proposed NMR based method has the following merits:

(1) *The nuclear norm is more robust than the Euclidean ($L_2$) norm as a similarity measure for alleviating the effect of illumination.* The current methods like LRC, SRC and CRC all use $L_2$ norm to define the decision rule. We will show that $L_2$ norm is sensitive to the extreme illumination changes, while the nuclear norm is not.

(2) *The low-rank assumption is more intuitive than the sparseness assumption for removing the structural noise (like the block occlusion).* It is natural that the rank of the occluded part in an image is lower than the rank of the whole image. When the size of the occluded part becomes larger, the noised pixels are not sparse, but the error image is still low-rank. On the other hand, in the sparse noise model like SRC, one needs to construct a big dictionary to represent noise, which causes a large computational burden. In contrast, our method does not need this dictionary and can recover the noise directly.

(3) *The proposed NMR does not need the assumption that the representation errors (or noises) of*

*pixels are statistically independent*. Note that this assumption is fundamental in the theoretical derivation of the spare robust regression model [6, 7]. Instead of working on pixel-errors individually, our model takes advantage of the low-rank structure of the error image and uses the nuclear norm to characterize it.

With respect to the use of the structural information of errors, in recent literature on face recognition, there are two papers deserving to be mentioned. Z. Zhou et al. incorporated the Markov Random Field model into the sparse representation framework for spatial continuity of the occlusion [11]. X. Li et al. explored the intrinsic structure of continuous occlusion and proposed the structured sparse error coding (SSEC) model [12]. The two methods share a same two-step iteration strategy: (1) Detecting error via sparse representation, and (2) Estimating error support (i.e. determining the real occluded part) using graph cuts. The difference is that SSEC uses more elaborate techniques such as the iteratively reweighted sparse coding in the error detection step and a morphological graph model in the error support step for achieving better performance. However, SSEC does not numerically converge to the desired solution; it needs an additional quality assessment model to choose the desired solution from the iteration sequence. Besides, SSEC contains many parameters, some of which are sensitive and have a significant effect on the performance. It can be said that our NMR provides a unified framework to integrate error detection and error support into one simple model. It just has one parameter, which is easily tuned and relatively insensitive to variations of databases.

The reminder of the paper is organized as follows: Section 2 introduced the nuclear norm based matrix regression problem and algorithm. In Section 3, we presented the NMR classifier for face recognition. In Section 4, we conducted experiments on four public face databases and compared with the state-of-the-art methods. Finally, Section 5 concludes the paper.

## 2. Nuclear Norm Based Matrix Regression

This section first formulates the nuclear norm based matrix regression (NMR) problem, then applies the alternating direction method of multipliers to solve the problem, and finally present the convergence analysis of the proposed algorithm.

**2.1 Problem Formulation**

Given a set of n image matrices $\mathbf{A}_1, \cdots, \mathbf{A}_n \in R^{p \times q}$ and an image matrix $\mathbf{B} \in R^{p \times q}$, let us represent $\mathbf{B}$ linearly using $\mathbf{A}_1, \cdots, \mathbf{A}_n$, i.e.

$$\mathbf{B} = x_1\mathbf{A}_1 + x_2\mathbf{A}_2 + \cdots + x_n\mathbf{A}_n + \mathbf{E}, \tag{1}$$

where $x_1, x_2, \cdots, x_n$ is a set of representation coefficients, and $\mathbf{E}$ is the representation residual.

Let us define the following linear mapping from $R^n$ to $R^{p \times q}$:

$$A(\mathbf{x}) = x_1\mathbf{A}_1 + x_2\mathbf{A}_2 + \cdots + x_n\mathbf{A}_n \tag{2}$$

Then, the formula (1) becomes

$$\mathbf{B} = A(\mathbf{x}) + \mathbf{E}, \tag{3}$$

The formula (3) or (1) gives a general form of a linear *matrix regression* model, in contrast with the classical linear *vector regression* model. If we use the Frobenius norm to evaluate the regression coefficients by

$$\min_x \|A(x) - \mathbf{B}\|_F^2 \tag{4}$$

The solution of the linear matrix regression is identical to the least-squares solution of the corresponding linear vector regression problem

$$Vec(\mathbf{B}) = x_1 Vec(\mathbf{A}_1) + \cdots + x_n Vec(\mathbf{A}_n) + \mathbf{E}, \tag{5}$$

where $Vec(\mathbf{B})$ is an operator converting the matrix $\mathbf{B}$ into a vector.

Motivated by observations or requirements that the residual image $A(x) - \mathbf{B}$ at the optimal solution is typically low rank in many applications, we would like to evaluate the regression coefficients via

$$\min_x \mathrm{rank}(A(x) - \mathbf{B}) \tag{6}$$

The above rank minimization problem is converted into the nuclear norm minimization problem for optimization in spirit of [18-25]:

$$\min_x \|A(x) - \mathbf{B}\|_* \tag{7}$$

Moreover, borrowing the idea of the Ridge regression, we would like to add a similar regularization term to Eq. (7) and obtain the regularized *matrix regression* model

$$\min_x \|A(\mathbf{x}) - \mathbf{B}\|_* + \tfrac{1}{2}\lambda \|\mathbf{x}\|_2^2 \tag{8}$$

We will discuss how to solve this model in the following subsection.

## 2.2 Algorithm

The alternating direction method of multipliers (ADMM) or the augmented Lagrange multipliers

(ALM) method has been applied to the nuclear norm optimization problems [28, 29]. For more details of ADMM, one can refer the paper [30]. Motivated by Hansson's work [29], we here provide details of using ADMM to solve the regularized matrix regression problem.

The model in (8) can be rewritten as

$$\min \| \mathbf{Y} \|_* + \tfrac{1}{2} \lambda \| \mathbf{x} \|_2^2 \quad subject\ to \quad A(\mathbf{x}) - \mathbf{B} = \mathbf{Y} \tag{9}$$

The augmented Lagrangian function $L_t$ is defined by

$$L_\mu(\mathbf{Y}, \mathbf{x}, \mathbf{Z}) = \| \mathbf{Y} \|_* + \tfrac{1}{2} \lambda \| \mathbf{x} \|_2^2 + \mathrm{Tr}\left(\mathbf{Z}^T \left(A(\mathbf{x}) - \mathbf{Y} - \mathbf{B}\right)\right) + \tfrac{\mu}{2} \| A(\mathbf{x}) - \mathbf{Y} - \mathbf{B} \|_F^2, \tag{10}$$

where $\mu > 0$ is a penalty parameter, $\mathbf{Z}$ is the Lagrange multipliers, and $\mathrm{Tr}(\cdot)$ is the trace operator. Note that if $\mu = 0$, the Eq. (10) is the standard Lagrangian function.

ADMM consists of the following iterations

(i) Given $\mathbf{Y} = \mathbf{Y}^k$ and $\mathbf{Z} = \mathbf{Z}^k$, updating $\mathbf{x}$ by

$$\mathbf{x}^{k+1} = \arg\min_x L_\mu(\mathbf{Y}, \mathbf{x}, \mathbf{Z}) \tag{11}$$

(ii) Given $\mathbf{x} = \mathbf{x}^{k+1}$ and $\mathbf{Z} = \mathbf{Z}^k$, updating $\mathbf{Y}$ by

$$\mathbf{Y}^{k+1} = \arg\min_\mathbf{Y} L_\mu(\mathbf{Y}, \mathbf{x}, \mathbf{Z}) \tag{12}$$

(iii) Given $\mathbf{x} = \mathbf{x}^{k+1}$ and $\mathbf{Y} = \mathbf{Y}^{k+1}$, Updating $\mathbf{Z}$ by

$$\mathbf{Z}^{k+1} = \mathbf{Z}^k + \mu\left(A(\mathbf{x}) - \mathbf{Y} - \mathbf{B}\right) \tag{13}$$

The key steps are to solve the optimization problems in Eqs. (11) and (12). For convenience, let us rewrite the augmented Lagrangian function in a different form. Since the last two items in Eq. (10) can be alternatively expressed as

$$\mathrm{Tr}\left(\mathbf{Z}^T\left(A(\mathbf{x}) - \mathbf{Y} - \mathbf{B}\right)\right) + \tfrac{\mu}{2} \| A(\mathbf{x}) - \mathbf{Y} - \mathbf{B} \|_F^2$$

$$= \mathrm{Tr}\left(\mathbf{Z}^T(A(\mathbf{x}) - \mathbf{Y} - \mathbf{B})\right) + \tfrac{\mu}{2} \mathrm{Tr}\left((A(\mathbf{x}) - \mathbf{B} - \mathbf{Y})^T(A(\mathbf{x}) - \mathbf{B} - \mathbf{Y})\right)$$

$$= \tfrac{\mu}{2} \mathrm{Tr}\left([A(\mathbf{x}) - (\mathbf{B} + \mathbf{Y} - \tfrac{1}{\mu}\mathbf{Z})]^T [A(\mathbf{x}) - (\mathbf{B} + \mathbf{Y} - \tfrac{1}{\mu}\mathbf{Z})]\right) - \tfrac{1}{2\mu} \mathrm{Tr}\left(\mathbf{Z}^T \mathbf{Z}\right)$$

$$= \tfrac{\mu}{2} \| A(\mathbf{x}) - (\mathbf{B} + \mathbf{Y} - \tfrac{1}{\mu}\mathbf{Z}) \|_F^2 - \tfrac{1}{2\mu} \| \mathbf{Z} \|_F^2, \tag{14}$$

we have

$$L_\mu(\mathbf{Y}, \mathbf{x}, \mathbf{Z}) = \| \mathbf{Y} \|_* + \tfrac{1}{2} \lambda \| \mathbf{x} \|_2^2 + \tfrac{\mu}{2} \| A(\mathbf{x}) - (\mathbf{B} + \mathbf{Y} - \tfrac{1}{\mu}\mathbf{Z}) \|_F^2 - \tfrac{1}{2\mu} \| \mathbf{Z} \|_F^2 \tag{15}$$

Based on the above form of the augmented Lagrangian function, it is easy to solve the problems in

Eqs. (11) and (12). Specifically, Eq. (11) can be expressed as

$$\mathbf{x}^{k+1} = \arg\min_{x} L_\mu(\mathbf{Y}, \mathbf{x}, \mathbf{Z})$$

$$= \arg\min_{x} \left( \tfrac{\mu}{2} \| \mathbf{A}(\mathbf{x}) - (\mathbf{B} + \mathbf{Y} - \tfrac{1}{\mu}\mathbf{Z}) \|_F^2 + \tfrac{1}{2}\lambda \| \mathbf{x} \|_2^2 \right) \quad (16)$$

Eq. (12) can be expressed as

$$\mathbf{Y}^{k+1} = \arg\min_{\mathbf{Y}} L_\mu(\mathbf{Y}, \mathbf{x}, \mathbf{Z})$$

$$= \arg\min_{\mathbf{Y}} \left( \| \mathbf{Y} \|_* + \tfrac{\mu}{2} \| \mathbf{A}(\mathbf{x}) - (\mathbf{B} + \mathbf{Y} - \tfrac{1}{\mu}\mathbf{Z}) \|_F^2 \right) \quad (17)$$

Now, we consider how to solve the problem in Eq. (16).

Letting $\mathbf{H} = [\text{Vec}(\mathbf{A}_1), \cdots, \text{Vec}(\mathbf{A}_n)]$, we can rewrite $\mathbf{A}(\mathbf{x}) = \sum_{j=1}^{n} x_j \mathbf{A}_j$ into the matrix form $\mathbf{Hx}$.

Denote $\mathbf{g} = \text{Vec}\left(\mathbf{B} + \mathbf{Y} - \tfrac{1}{\mu}\mathbf{Z}\right)$. Therefore, Eq. (16) is equivalent to

$$\mathbf{x}^{k+1} = \arg\min_{x} \left( \tfrac{\mu}{2} \| \mathbf{Hx} - \mathbf{g} \|_2^2 + \tfrac{1}{2}\lambda \| \mathbf{x} \|_2^2 \right)$$

$$= \arg\min_{x} \left( \| \mathbf{Hx} - \mathbf{g} \|_2^2 + \tfrac{\lambda}{\mu} \| \mathbf{x} \|_2^2 \right) \quad (18)$$

Since Eq. (18) is a standard Ridge regression model, we can get its close-form solution

$$\mathbf{x}^{k+1} = (\mathbf{H}^T \mathbf{H} + \tfrac{\lambda}{\mu} \mathbf{I})^{-1} \mathbf{H}^T \mathbf{g} \quad (19)$$

In the following, we consider how to solve the problem in Eq. (17). Eq. (17) is equivalent to

$$\mathbf{Y}^{k+1} = \arg\min_{\mathbf{Y}} \left( \tfrac{1}{\mu} \| \mathbf{Y} \|_* + \tfrac{1}{2} \| \mathbf{Y} - (\mathbf{A}(\mathbf{x}) - \mathbf{B} + \tfrac{1}{\mu}\mathbf{Z}) \|_F^2 \right) \quad (20)$$

The optimal solution can be computed via the singular value thresholding algorithm [31]. Specifically, given a matrix $\mathbf{Q} \in R^{p \times q}$ of rank $r$, the singular value decomposition (SVD) of X is

$$\mathbf{Q} = \mathbf{U}_{p \times r} \Sigma \mathbf{V}_{q \times r}^T, \quad \Sigma = diag(\sigma_1, \cdots, \sigma_r), \quad (21)$$

where $\sigma_1, \cdots, \sigma_r$ are positive singular values, and $\mathbf{U}_{p \times r}$ and $\mathbf{V}_{q \times r}$ are corresponding matrices with orthogonal columns.

For a given $\tau > 0$, the singular value shrinkage operator is defined as follows

$$D_\tau(\mathbf{Q}) = \mathbf{U}_{p \times r} \text{diag}\left( \{\max(0, \sigma_j - \tau)\}_{1 \le j \le r} \right) \mathbf{V}_{q \times r}^T \quad (22)$$

**Theorem 1** [31] For each $\mathbf{X} \in R^{p \times q}$ and $\tau > 0$, the singular value shrinkage operator in (22) obeys

$$D_\tau(\mathbf{Q}) = \arg\min_{\mathbf{Y}} \left( \tau \| \mathbf{Y} \|_* + \tfrac{1}{2} \| \mathbf{Y} - \mathbf{Q} \|_F^2 \right) \quad (23)$$

From Theorem 1, the optimal solution of (20) is

$$\mathbf{Y} = D_{\frac{1}{\mu}}\left(\mathbf{A}(\mathbf{x}) - \mathbf{B} + \frac{1}{\mu}\mathbf{Z}\right) \tag{24}$$

In summary, the core of ADMM algorithm for nuclear norm based matrix regression (NMR) problem involves two sub-problems: the ridge regression and the singular value thresholding. The sketch of our algorithm is given in Fig. 2.

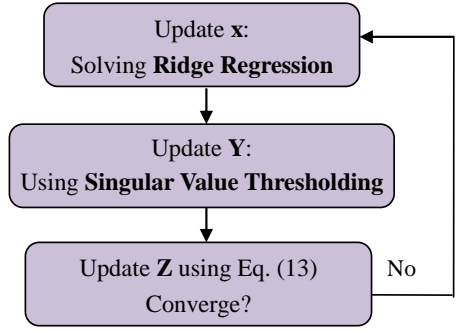

**Fig.2** The sketch of ADMM algorithm for NMR

Boyd et al. [30] give the optimality conditions and stopping criteria of the ADMM algorithm. Based on the results in [29, 30], we use the following termination criterion: the primal and dual residuals must be small, i.e.

$$\|\mathbf{r}_{pri}^k\|_2 \leq \varepsilon_{pri} \quad \text{and} \quad \|\mathbf{s}_{dual}^k\|_2 \leq \varepsilon_{dual}, \tag{25}$$

where $\mathbf{r}_{pri}^k$, $\mathbf{s}_{dual}^k$, $\varepsilon_{pri}$, and $\varepsilon_{dual}$ are defined as follows

$$\mathbf{r}_{pri}^k = \mathbf{A}(\mathbf{x}^k) - \mathbf{Y}^k - \mathbf{B}, \tag{26}$$

$$\mathbf{s}_{dual}^k = \mu \mathbf{H}^T \operatorname{Vec}(\mathbf{Y}^k - \mathbf{Y}^{k-1}) \tag{27}$$

$$\varepsilon_{pri} = \sqrt{pq}\varepsilon_{abs} + \varepsilon_{rel} \max\{\|\mathbf{A}(\mathbf{x})\|_F, \|\mathbf{Y}\|_F, \|\mathbf{B}\|\} \tag{28}$$

$$\varepsilon_{dual} = \sqrt{n}\varepsilon_{abs} + \varepsilon_{rel} \|\mathbf{H}^T \operatorname{Vec}(\mathbf{Z})\|_2 \tag{29}$$

Finally, we would like to further elaborate our algorithm. If we fix penalty parameter $\mu$ in the augmented Lagrangian function, the updating $\mathbf{x}$ step via solving ridge regression problem can be computed more efficiently. Looking back at the Eq. (19), $(\mathbf{H}^T\mathbf{H} + \frac{\lambda}{\mu}\mathbf{I})^{-1}\mathbf{H}^T$ is fixed in each iteration, so it can be calculated and saved in advance. Let

$$\mathbf{M} = (\mathbf{H}^T\mathbf{H} + \frac{\lambda}{\mu}\mathbf{I})^{-1}\mathbf{H}^T. \tag{30}$$

Then, in each iteration for updating $\mathbf{x}$, we only need to update $\mathbf{g} = \operatorname{Vec}\left(\mathbf{B} + \mathbf{Y} - \frac{1}{\mu}\mathbf{Z}\right)$ and carry out the matrix multiplication $\mathbf{Mg}$ once. On the other hand, in the iteration for updating $\mathbf{Y}$, the main computation is consumed for performing the singular value decomposition of the matrix $\mathbf{Q} = \mathbf{A}(\mathbf{x}) - \mathbf{B} + \frac{1}{\mu}\mathbf{Z}$. Since $\mathbf{Q}$ has the same size as the image matrix, the computational complexity of this step only depends on the size of images.

The detailed ADMM algorithm for NMR is summarized in Algorithm 1.

---
**Algorithm 1** NMR Algorithm via ADMM

**Input:** A set of image matrices $\mathbf{A}_1, \cdots, \mathbf{A}_n$ and an image matrix $\mathbf{B} \in R^{p \times q}$, the model parameters $\lambda$ and $\mu$, the termination condition parameters $\varepsilon_{abs}$ and $\varepsilon_{rel}$.

---
1: Let $\mathbf{H} = [\text{Vec}(\mathbf{A}_1), \cdots, \text{Vec}(\mathbf{A}_n)]$. Compute
   $\mathbf{M} = (\mathbf{H}^T\mathbf{H} + \frac{\lambda}{\mu}\mathbf{I})^{-1}\mathbf{H}^T$;
2: $\mathbf{Y}^0 = -\mathbf{B}$, k = 0;
3: Updating $\mathbf{x}$: Let $\mathbf{g} = \text{Vec}(\mathbf{B} + \mathbf{Y}^k - \frac{1}{\mu}\mathbf{Z}^k)$. $\mathbf{x}^{k+1} = \mathbf{Mg}$;
4: Updating $\mathbf{Y}$: $\mathbf{Y}^{k+1} = D_{\frac{1}{\mu}}\left(\mathbf{A}(\mathbf{x}^{k+1}) - \mathbf{B} + \frac{1}{\mu}\mathbf{Z}^k\right)$;
5: Updating $\mathbf{Z}$: $\mathbf{Z}^{k+1} = \mathbf{Z}^k + \mu\left(\mathbf{A}(\mathbf{x}^{k+1}) - \mathbf{Y}^{k+1} - \mathbf{B}\right)$;
6. If Eq. (25) is not satisfied go to Step 3.

**Output:** Optimal regression coefficient vector $\mathbf{x}^{k+1}$

---

Algorithm 1 can be interpreted in the two-step iteration strategy for robust face recognition as that adopted in [11, 12]. Step 3 updating x is actually an error detection step for determining the representation coefficients and representation errors, and Step 4 updating Y is actually an error support step for determining the real corrupted part. So, we can say that NMR provides a unified framework to integrate error detection and error support into one simple model.

*The computational complexity of the NMR Algorithm*

Given the training sample size *n* and the image size $p \times q$, and let $m = p \times q$. The computational complexity of Step 3 is $O(mn)$, which is determined by the matrix multiplication $\mathbf{Mg}$. The computational complexity of Step 4 is $O(\min(p^2q, pq^2))$, which is determined by the singular value decomposition of a $p \times q$ matrix $\mathbf{Q} = \mathbf{A}(\mathbf{x}) - \mathbf{B} + \frac{1}{\mu}\mathbf{Z}$. In the case that *p=q*, the computational complexity becomes $O(m^{1.5})$. So, the computational complexity of the NMR Algorithm is $O(k(m^{1.5} + mn))$, where *k* is the number of iterations. Empirically, our algorithm stops within 20 iterations, i.e. $k<20$.

## 2.3 Convergence Analysis

In this section, we will give a convergence analysis of ADMM for computing the regression coefficient vector. In order to guarantee the convergence, some assumptions are very important, which are supposed to be simple and easily satisfied. Furthermore, if simple assumptions can guarantee strong convergence results, this algorithm will be more applicable. Indeed, Algorithm 1 is a special case of a more general class of augmented Lagrange multiplier algorithms known as alternating directions methods [40]. The convergence of these algorithms has been studied extensively (see [41, 42] and many references therein, as well as discussions in [28, 40]). In recent years, the existence of the saddle points is a popular assumption for the convergence of algorithms. For example, S. Boyd et al. investigated convergence of ADMM by virtue of the properties of the saddle points, and give three important results: Residual convergence, Objective convergence and Dual variable convergence [33]. However, the objective convergence cannot deduce the optimal point of iterative process. If the optimal point of iterative process could be pointed out, the iterative trend will be clearer. Thus, we here mainly study the accumulation points of the iterative variables for Algorithm 1.

Let $(\mathbf{Y}^*, \mathbf{x}^*, \mathbf{Z}^*)$ be a saddle point of the following Lagrangian function: $L(\mathbf{Y}, \mathbf{x}, \mathbf{Z}) = \|\mathbf{Y}\|_{S_p}^p + \frac{1}{2}\lambda \|\mathbf{x}\|_2^2 + \mathrm{Tr}(\mathbf{Z}^T(\mathbf{A}(\mathbf{x}) - \mathbf{Y} - \mathbf{B}))$, $q^k = \|\mathbf{Y}^k\|_* + \frac{\lambda}{2}\|\mathbf{x}^k\|_2^2$, $q^* = \|\mathbf{Y}^*\|_* + \frac{\lambda}{2}\|\mathbf{x}^*\|_2^2$ and $\mathbf{R}^k = \mathbf{A}(\mathbf{x}^k) - \mathbf{Y}^k - \mathbf{B}$, $\mathbf{r}^k = vec(\mathbf{R}^k)$. According to the analysis in [33, 43], finding the optimal solutions of original and dual problems is equivalent to finding a saddle point of the function $L$. Thus, $\mathbf{Z}^*$ is dual optimal. In addition, from [44], we can know that

$$\mathbf{Z}^k \to \mathbf{Z}^*, \text{ as } k \to \infty. \tag{31}$$

**Theorem 2.** If $\mu > 0$, then the sequence $\{(\mathbf{Y}^k, \mathbf{x}^k, \mathbf{Z}^k)\}$ generated by Algorithm 1 converges to a saddle point $(\mathbf{Y}^*, \mathbf{x}^*, \mathbf{Z}^*)$ of the Lagrangian function $L$.

**Proof.** By the definition of the saddle point, we have $L(\mathbf{Y}^*, \mathbf{x}^*, \mathbf{Z}^*) \leq L(\mathbf{Y}^k, \mathbf{x}^k, \mathbf{Z}^*)$, which can be written as

$$q^* - q^{k+1} \leq tr(\mathbf{Z}^{*T}\mathbf{R}^{k+1}). \tag{32}$$

Meanwhile, from $\mathbf{x}^{k+1} = \arg\min_x L_\mu(\mathbf{Y}^k, \mathbf{x}, \mathbf{Z}^k)$, we have

$$0 = \partial L_\mu(\mathbf{Y}^k, \mathbf{x}^{k+1}, \mathbf{Z}^k) = \lambda \mathbf{x}^{k+1} + \mu \mathbf{H}^T \left( \mathbf{H}\mathbf{x}^{k+1} - Vec\left( \mathbf{B} + \mathbf{Y}^k - \frac{1}{\mu}\mathbf{Z}^k \right) \right)$$
$$= \lambda \mathbf{x}^{k+1} + \mathbf{H}^T \left( Vec(\mathbf{Z}^{k+1}) + \mu Vec(\mathbf{Y}^{k+1} - \mathbf{Y}^k) \right).$$

This implies that $\mathbf{x}^{k+1}$ minimizes $\frac{\lambda}{2}\|\mathbf{x}\|_2^2 + (Vec(\mathbf{Z}^{k+1}) - \mu Vec(\mathbf{Y}^{k+1} - \mathbf{Y}^k))^T \mathbf{H}\mathbf{x}$. A similar derivation leads to that $\mathbf{Y}^{k+1}$ minimizes $\|\mathbf{Y}\|_* - Tr((\mathbf{Z}^{k+1})^T \mathbf{Y})$. Therefore,

$$\frac{\lambda}{2}\|\mathbf{x}^{k+1}\|_2^2 + (Vec(\mathbf{Z}^{k+1}) - \mu Vec(\mathbf{Y}^{k+1} - \mathbf{Y}^k))^T \mathbf{H}\mathbf{x}^{k+1} \leq \frac{\lambda}{2}\|\mathbf{x}^*\|_2^2 + (Vec(\mathbf{Z}^{k+1}) - \mu Vec(\mathbf{Y}^{k+1} - \mathbf{Y}^k))^T A\mathbf{x}^*,$$

$$\|\mathbf{Y}^{k+1}\|_* - Tr((\mathbf{Z}^{k+1})^T \mathbf{Y}^{k+1}) \leq \|\mathbf{Y}^*\|_* - Tr((\mathbf{Z}^{k+1})^T \mathbf{Y}^*).$$

Adding the above inequalities by virtue of $\mathbf{H}\mathbf{x}^{k+1} - \mathbf{H}\mathbf{x}^* = \mathbf{r}^{k+1} + vec(\mathbf{Y}^{k+1} - \mathbf{Y}^*)$ and $q^* = \|\mathbf{Y}^*\|_* + \frac{\lambda}{2}\|\mathbf{x}^*\|_2^2$, and after rearranging, we have

$$q^{k+1} - q^* \leq -Vec(\mathbf{Z}^k) \cdot \mathbf{r}^k \mu - Vec(\mathbf{Y}^{k} - \mathbf{Y}^k)(\cdot Vec^k \mathbf{Y}^{+k}). \tag{33}$$

Adding (32) and (33), regrouping terms, and multiplying through by 2 results in

$$2Vec^T(\mathbf{Z}^{k+1} - \mathbf{Z}^*) \cdot \mathbf{r}^{k+1} + 2\mu Vec(\mathbf{Z}^{+k}\mathbf{Z}) \cdot \mathbf{r}^{+1k} 2\mu (\mathbf{Z}^T - \mathbf{Z}^k)(\mathbf{Y}^{+k} - \mathbf{Y}^*) \leq \tag{34}$$

By the known conditions in Algorithm 1 and some algebraic techniques, we can convert (34) to the following form:

$$\left[ (1/\mu)\|Vec^T(\mathbf{Z}^k - \mathbf{Z}^*)\|_2^2 + \mu\|\mathbf{Y}^k - \mathbf{Y}^*\|_F^2 \right] - \left[ (1/\mu)\|Vec^T(\mathbf{Z}^{k+1} - \mathbf{Z}^*)\|_2^2 + \mu\|\mathbf{Y}^{k+1} - \mathbf{Y}^*\|_F^2 \right]$$
$$\geq \mu\|\mathbf{R}^{k+1} + (\mathbf{Y}^{k+1} - \mathbf{Y}^k)\|_F^2 = \mu\|\mathbf{r}^{k+1}\|_2^2 + \mu\|\mathbf{Y}^{k+1} - \mathbf{Y}^k\|_F^2 + 2\mu\mathbf{r}^{k+1} \cdot Vec(\mathbf{Y}^{k+1} - \mathbf{Y}^k). \tag{35}$$

Evidently, $2\mu\mathbf{r}^{k+1} \cdot Vec(\mathbf{Y}^{k+1} - \mathbf{Y}^k) \geq 0$, thus, (35) can be written as

$$\left[ (1/\mu)\|\mathbf{Z}^k - \mathbf{Z}^*\|_F^2 + \mu\|\mathbf{Y}^k - \mathbf{Y}^*\|_F^2 \right] - \left[ (1/\mu)\|\mathbf{Z}^{k+1} - \mathbf{Z}^*\|_F^2 + \mu\|\mathbf{Y}^{k+1} - \mathbf{Y}^*\|_F^2 \right]$$
$$\geq \mu\|\mathbf{R}^{k+1}\|_F^2 + \mu\|\mathbf{Y}^{k+1} - \mathbf{Y}^k\|_F^2. \tag{36}$$

Let $U^k = (1/\mu)\|\mathbf{Z}^k - \mathbf{Z}^*\|_F^2 + \mu\|\mathbf{Y}^k - \mathbf{Y}^*\|_F^2$, then (36) states that $U^{k+1} \leq U^k \leq U^0$. It is easy to see that $\mathbf{Z}^k$ and $\mathbf{Y}^k$ are bounded. Iterating the inequality (36) gives that $\mu\sum_{k=0}^{\infty}\|\mathbf{R}^{k+1}\|_F^2 + \mu\|\mathbf{Y}^{k+1} - \mathbf{Y}^k\|_F^2 \leq U^0$, which implies that $\mathbf{R}^k \to 0$ and $\mathbf{Y}^{k+1} - \mathbf{Y}^k \to 0$ as $k \to \infty$.

Thus, $-tr(\mathbf{Z}^{*T}\mathbf{R}^{k+1}) \to 0$ and $-Vec^T(\mathbf{Z}^{k+1}) \cdot \mathbf{r}^k - \mu Vec^T(\mathbf{Y}^{k+1} - \mathbf{Y}^k) \cdot Vec(\mathbf{r}^{k+1} + \mathbf{Y}^{k+1} - \mathbf{Y}^*) \to 0$ as $k \to \infty$.

From (32) and (33), we have $\lim_{k \to \infty} q^k = q^*$. Therefore, there exists an increasing subsequence $\{q^{k_j}\}$ of $\{q^k\}$ such that $\lim_{j \to \infty} q^{k_j} = q^*$.

That is to say, there exists a subsequence $\{\mathbf{Y}^{k_j}\}$ of $\{\mathbf{Y}^k\}$ and a subsequence $\{\mathbf{x}^{k_j}\}$ of $\{\mathbf{x}^k\}$ such that $\lim_{j\to\infty}\|\mathbf{Y}^{k_j}\|_* + \frac{\lambda}{2}\|\mathbf{x}^{k_j}\|_2^2 = q^*$ and $\|\mathbf{Y}^*\|_* + \frac{\lambda}{2}\|\mathbf{x}^*\|_2^2 \geq \|\mathbf{Y}^{k_{j+1}}\|_* + \frac{\lambda}{2}\|\mathbf{x}^{k_{j+1}}\|_2^2 \geq \|\mathbf{Y}^{k_j}\|_* + \frac{\lambda}{2}\|\mathbf{x}^{k_j}\|_2^2$, which means that $(\mathbf{Y}_{k_j}, \mathbf{x}_{k_j})$ approaches to $(\mathbf{Y}^*, \mathbf{x}^*)$ as $j \to \infty$.

Since $\mathbf{Z}^k \to \mathbf{Z}^*$, we have $U^{k_j} = \left[(1/\mu)\|\mathbf{Z}^{k_j} - \mathbf{Z}^*\|_F^2 + \mu\|\mathbf{Y}^{k_j} - \mathbf{Y}^*\|_F^2\right] \to 0$. From $U^{k+1} \leq U^k \leq U^0$, we get that $U^k = \left[(1/\mu)\|\mathbf{Z}^k - \mathbf{Z}^*\|_F^2 + \mu\|\mathbf{Y}^k - \mathbf{Y}^*\|_F^2\right] \to 0$. Thus, $\mathbf{Y}^k \to \mathbf{Y}^*$. By $\lim_{k\to\infty} q^k = q^*$, we know that $\mathbf{x}^k \to \mathbf{x}^*$.

## 3. NMR based Classification

This section first shows the similarity and occlusion observations. Based on these observations, the NMR based classifier is presented.

### 3.1 Observations

A classifier must involve a similarity measure (or distance) between the test sample and the class training samples. For achieving good performance, the distance between the test sample and the class that it belongs to should be smaller than the distance between the test sample and the class that it does not belong to. In this paper, we use the nuclear norm of the representation residual image to characterize the distance between test samples and classes. We will show the rationale of this distance characterization.

In many image recognition problems like face recognition, illumination changes make an important effect on the performance. So, a robust classifier (or similarity measure) should be insensitive to the illumination changes. We will show that the nuclear norm is less sensitive to the changes of lighting conditions than the Euclidean norm. On the other hand, the occlusion is another critical factor for image recognition. A robust classifier must have the ability of removing or alleviating the effect of occlusions. NMR provides a mechanism for dealing with occlusion in test images.

The NMR based classifier is motivated by the following two observations:

*(I) Similarity Observations*

If one uses the class sample images to represent a test image, the homo-class representation leads to a *low-rank residual image*. In other words, *the nuclear norm of residual image is minimal if and only if the test sample is represented by samples of the same class*. This fact can be observed in the following example.

Given two classes of face images selected from the extended Yale B database, as shown in Fig. 3 (a), the first row of images belong to Class I, and the second row of images belong to Class II. In this example, $\mathbf{A}_{ij}(i=1,2; j=1,\cdots,6)$ are used for training, and $\mathbf{B}_i(i=1,2)$ for test.

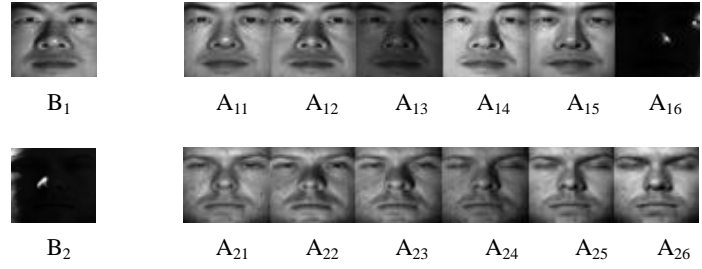

(a) Testing and Training images of two classes of faces

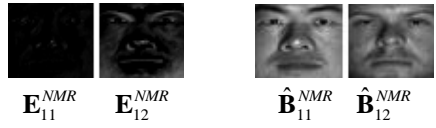

(b) The residual images (left two) and reconstructed images (right two) of $B_1$ using NMR

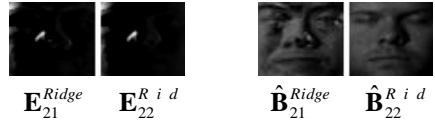

(c) The residual images (left two) and reconstructed images (right two) of $B_2$ using Ridge regression

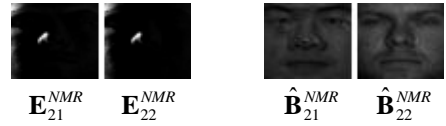

(d) The residual images (left two) and reconstructed images (right two) of $B_2$ using NMR

Fig. 3 Example for similarity observations

We first represent the image $B_1$ using the two classes of training samples via NMR. The resulting residual images and the reconstructed images are shown in Fig. 3 (b), where $E_{11}$ is the homo-class representation residual, and $E_{12}$ is the hetero-class representation residual. $\|\mathbf{E}_{11}^{NMR}\|_* = 6.76$ is much smaller than $\|\mathbf{E}_{12}^{NMR}\|_* = 21.76$. So, the nuclear norm of class residual image can be used as a similarity measure to design the rule for classification.

In the following, we will show why we use the nuclear norm instead of the Euclidean norm. The Euclidean norm, as a similarity measure, has been widely used in most regression methods based classifiers, like the LRC [1], SRC [4], and CRC [16]. However, the Euclidean norm is sensitive to illumination changes. For example, let us consider the test image $B_2$, which belongs to Class II but with quite different lighting conditions from the training samples of Class II. Note that there is one image in training set of Class I, $A_{16}$, which has the similar lighting condition as $B_2$.

We first represent the image $B_2$ using the two classes of training samples via the Ridge regression (with Euclidean measure). The resulting residual images and reconstructed images are shown in Fig. 3 (c). We then represent $B_2$ using NMR and provide the residual images and reconstructed images in Fig. 3 (d). By comparing these figures, we find that the reconstructed images of $B_2$ using NMR look much better than those using linear Ridge regression. From this, we can deduce that the residual images resulting from the two methods should be different, although they look similar. In fact, these representation errors yield different results. For the Ridge regression, the homo-class representation error is larger than the hetero-class representation error, i.e.

$$\| \mathbf{E}_{22}^{Ridge} \|_2 = 4.54 > \| \mathbf{E}_{21}^{Ridge} \|_2 = 4.46 \tag{31}$$

That means the Ridge regression based classifier will misclassify the image $B_2$. However, for NMR, the homo-class representation error is still smaller than the hetero-class representation error, i.e.

$$\| \mathbf{E}_{22}^{NMR} \|_* = 11.02 < \| \mathbf{E}_{21}^{NMR} \|_* = 11.34 \tag{32}$$

That is, for the test image with such an extreme illumination condition, the nuclear norm of the homo-class representation residual image is still minimal. This motivates us to use NMR for classifier design.

*(II) Occlusion Observations*

It is intuitive to assume that the representation error image of an occluded image is low rank, since any occlusion only affects a part of image. We will show that if one uses clean training sample images to represent a test image with occlusion, the occluded part can be recovered by solving the NMR problem.

In our example, two classes of face images (with a resolution of 50x40) from the AR database, as shown Fig. 4(a), are used for training. We test two kinds of occluded images: the image with artificial occlusion, and the image with real-world occlusion. In Fig. 4 (b)-(e), the left column contains the occluded images. OI-1 and OI-2 are artificially generated. Specifically, OI-1 is generated by replacing a randomly-chosen part with a white square: a 25x25 matrix with all one elements, while OI-2 is generated in a similar way but with a larger white rectangle: a 35x25 matrix with all one elements.

In our tests, we use four methods to deal with occlusion. The first is our NMR method, the second is the Ridge regression, the third is the sparse representation (SR) model [4], and the fourth is the robust regression [5]. For each occluded image, the reconstructed images (recovered clean image) and the residual images (the recovered occlusion) are shown in Fig. 4 (b)-(e).

From Fig. 4 (b), it can be seen that for the 25x25 block occlusion (occluded pixels take 31.25% of all pixels), NMR perform as well as sparse representation (SR) for recovering the clean face and the occluded part. Both methods achieve better results than the Ridge regression and robust regression. However, Fig. 4 (c) shows that for the occlusion of a larger block size, 35x25, i.e., the occlusion rate up to 43.75%, NMR can still recover the clean face image well. In contrast, the other methods, including SR, fail to achieve good results.

Fig. 4 (d) and (e) show the recovered results of the four methods in the case of real-world occlusion. OI-3 is a face image with sunglasses (occlusion rate is relatively low), and OI-4 with a scarf (occlusion rate is relatively high). We can see that for the image with sunglass, NMR achieves comparable results with SR and robust regression, but for the image with a scarf, NMR significantly outperforms the other methods in recovering the clean image.

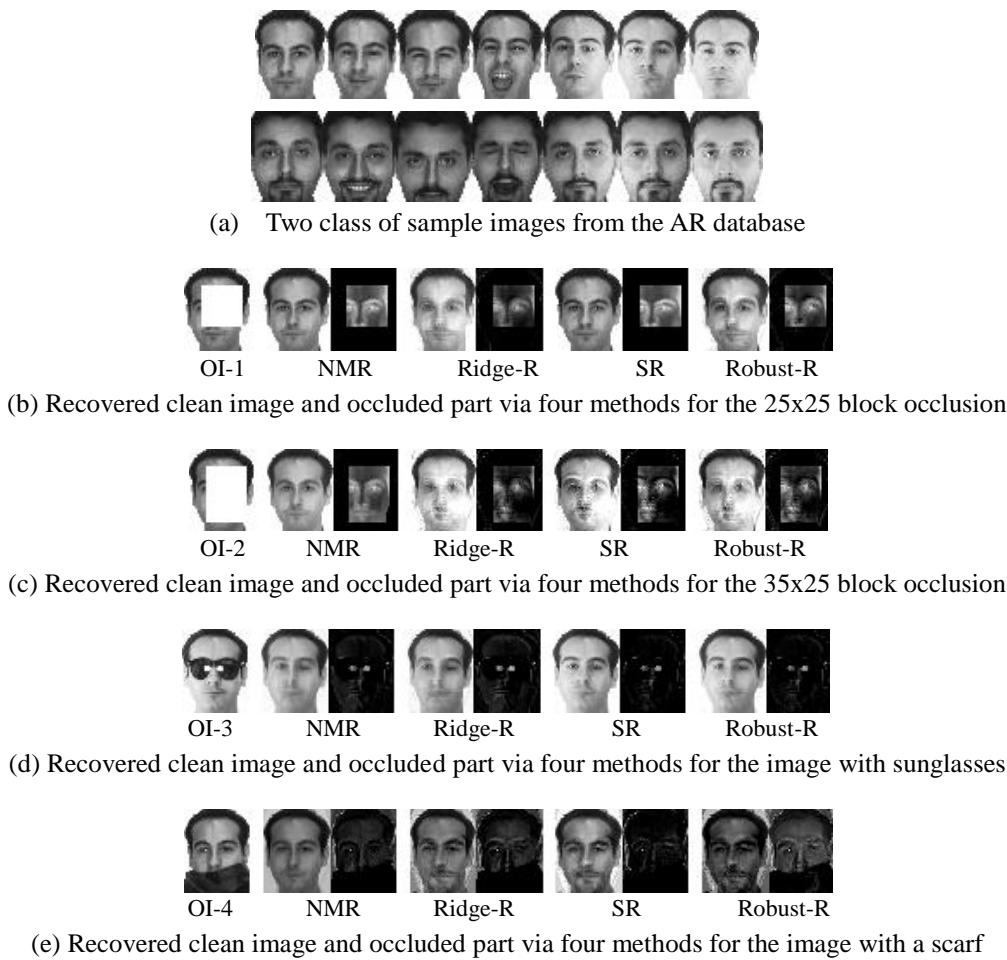

(a)   Two class of sample images from the AR database

OI-1     NMR     Ridge-R     SR     Robust-R
(b) Recovered clean image and occluded part via four methods for the 25x25 block occlusion

OI-2     NMR     Ridge-R     SR     Robust-R
(c) Recovered clean image and occluded part via four methods for the 35x25 block occlusion

OI-3     NMR     Ridge-R     SR     Robust-R
(d) Recovered clean image and occluded part via four methods for the image with sunglasses

OI-4     NMR     Ridge-R     SR     Robust-R
(e) Recovered clean image and occluded part via four methods for the image with a scarf

Fig. 4 Example for occlusion observations

## 3.2 NMR Classifier

Based on the above observations, we design our NRM classifier. Similar to the strategy of SRC, we use the training samples of all classes to form the set of regressors. Let $\mathbf{A}_1,\cdots,\mathbf{A}_n$ be training sample images of all classes. For a given test image $\mathbf{B}$, we use all training samples to represent it and obtain the representation coefficient vector by solving the following NMR model

$$\mathbf{x}^* = \arg\min_x \|A(\mathbf{x}) - \mathbf{B}\|_* + \tfrac{1}{2}\lambda \|\mathbf{x}\|_2^2 \qquad (33)$$

Based on the optimal solution $\mathbf{x}^*$, we get the reconstructed image of $\mathbf{B}$ as $\hat{\mathbf{B}} = A(\mathbf{x}^*)$, and the residual image $\mathbf{E} = \mathbf{B} - \hat{\mathbf{B}}$. If $\mathbf{B}$ is with occlusion, $\mathbf{E}$ reveals the recovered, occluded part according to the Occlusion Assumption.

Let $\delta_i : R^n \to R^n$ be the characteristic function that selects the coefficients associated with the $i$-th class. For $\mathbf{x} \in R^n$, $\delta_i(\mathbf{x})$ is a vector whose only nonzero entries are the entries in $\mathbf{x}$ that are associated with Class $i$. Using the coefficients associated with the i-th class, one can get the reconstruction of $\mathbf{B}$ in Class $i$ as $\hat{\mathbf{B}}_i = A(\delta_i(\mathbf{x}^*))$. The corresponding class reconstruction error is defined by

$$e_i(\mathbf{B}) = \|\hat{\mathbf{B}} - \hat{\mathbf{B}}_i\|_* = \left\|A(\mathbf{x}^*) - A(\delta_i(\mathbf{x}^*))\right\|_* \qquad (34)$$

The decision rule is defined as: if $e_l(\mathbf{B}) = \min_i e_i(\mathbf{B})$, then $\mathbf{B}$ is assigned to Class $l$.

## 4. Experiments

Four publicly available databases, the Extended Yale B database [35], the AR database [36], the Multi-PIE database [37] and the FRGC Database [38], are used in our experiments. The proposed method is tested and compared with state-of-the-art linear representation related classifiers: LRC [1], SRC [4], CRC [16], CESR[1] [9], RSC[2] [6], SSEC [12], HQ_A[3] and HQ_M[3] [34]. LRC and CRC are tuned to achieve their best performance by choosing the optimal regression parameters, and the parameter settings of the other methods follow the authors' suggestions. The default regression parameter of the proposed NMR is 1. It should be mentioned that here all experiments are done on the original face images, without any image preprocessing and feature extraction step.

---

[1] The Matlab source code: http://www.openpr.org.cn/index.php/All/69-CESR/View-details.html.
[2] The Matlab source code: http://www4.comp.polyu.edu.hk/~cslzhang/
[3] http://www.openpr.org.cn/index.php/91-Half-quadratic-based-Iterative-Minimization-for-Robust-Sparse-Representation/View-details.html

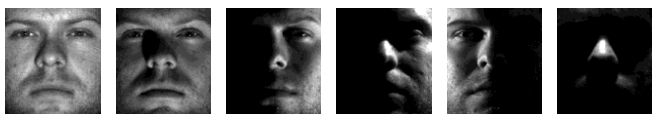

Fig. 5 Samples of a person under different illumination conditions in the Extended Yale B face database

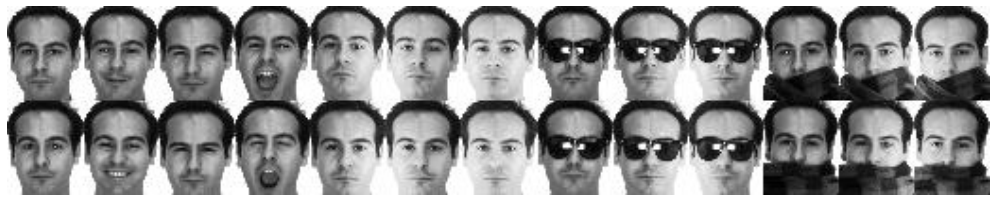

Fig. 6 Sample images of a person in the AR face database

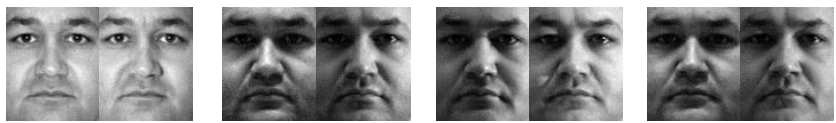

     Session 1        Session 2        Session 3        Session 4

Fig. 7 Sample images of a person under different illumination conditions in the Multi-PIE database from different sessions

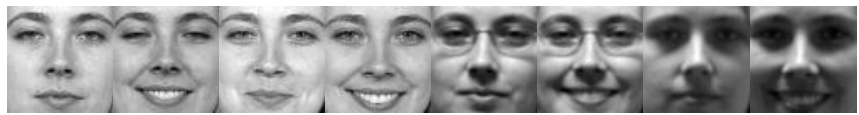

Fig. 8 Example FRGC images that have been cropped

### 4.1 Datasets

The involved four databases are described as follows:

***Extended Yale B***. The extended Yale B face database contains 38 persons under 9 poses and 64 illumination conditions [35]. The 64 images of a person in a particular pose are acquired at camera frame rate of 30 frames/ second, so there is only small change in head pose and facial expression for those 64 images. All frontal-face images marked with P00 are used, and each image is resized to 96×84 pixels in our experiment. Some sample images of one person are shown in Fig. 5.

***AR***. The AR face database contains over 4,000 color face images of 126 people (70 men and 56 women), including frontal views of faces with different facial expressions, lighting conditions and occlusions [36]. The pictures of most persons were taken in two sessions (separated by two weeks). Each section contains 13 color images and 120 individuals (65 men and 55 women) participated in both sessions. The images of these 120 individuals were selected and used in our experiment. We manually cropped the face portion of the image and then normalized it to 50×40 pixels. The normalized images of one person are shown in Fig. 6.

*Multi-PIE*. The CMU Multi-PIE database contains images of 337 different subjects with variations in pose, expression and illumination [37]. Individual attendance varies from 249, 203, 230 and 239 for Sessions 1-4. In our experiment, we use the frontal images with difficult illuminations and neutral expression. We manually cropped the face portion of the image and then normalized it to 50×40 pixels. The sample images of one person are shown in Fig. 7.

*FRGC*. The Face Recognition Grand Challenge (FRGC) version 2 database [38] contains 12,776 training images, 16,028 controlled target images, and 8,014 uncontrolled query images for the FRGC Experiment 4. The controlled images have good image quality, while the uncontrolled images display poor image quality, such as large illumination variations, low resolution of the face region, and possible blurring. It is these uncontrolled factors that pose the grand challenge to face recognition performance. Fig. 8 shows some examples of the FRGC dataset.

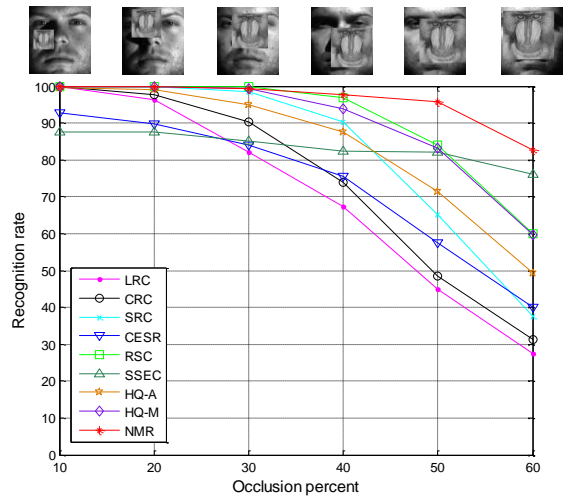

Fig. 9 Recognition rates (%) of LRC, CRC, SRC, CESR, RSC, SSEC, HQ_A, HQ_M and NMR under different levels of occlusion

## 4.2 Recognition with Random Occlusions

In the first experiment, we use the similar experiment setting as in [4] to test the performance of the proposed model. Subsets 1 and 2 of the Extended Yale B are used for training and Subset 3 for test. Each test image is corrupted by a randomly located square block of "baboon" image with varying block sizes. The block size determines the occlusion level of an image. The images on the top of Fig. 9 illustrate the occlusion levels varying from 10% to 60%.

Fig. 9 shows the recognition rates of LRC, CRC, SRC, CESR, RSC, SSEC, HQ_A (additive form),

HQ_M (Multiplicative form) and NMR under different occlusion levels. From Fig. 9, we can see that the proposed NMR significantly outperforms other robust methods such as SRC, RSC, HQ_M and SSEC, when the occlusion level is equal to or larger than 50%. When occlusion level is no more than 30%, SRC, RSC and HQ_M achieve similar results with NMR. The performance of SSEC is good when the occlusion level becomes high, but it has no advantage when the occlusion level is relative low. The recognition rates of LRC and CRC drop fast with the increase of occlusion levels, thus the two methods are sensitive to the level of structural noise.

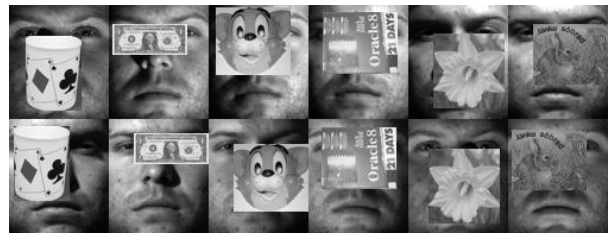

Fig. 10 Sample images of one person with different occlusions

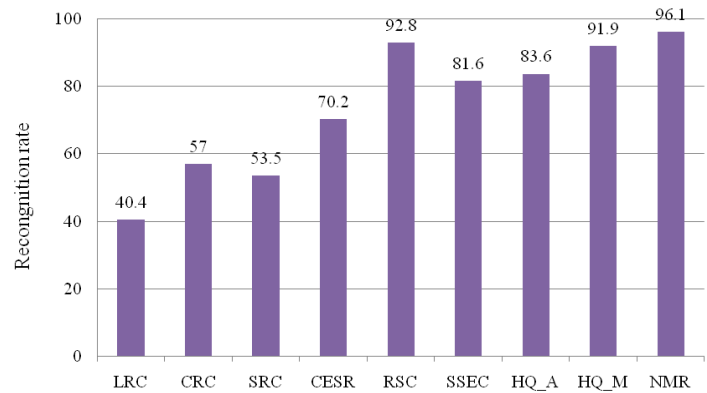

Fig. 11 Recognition rates (%) of LRC, CRC, SRC, CESR, RSC, RSC, SSEC, HQ_A, HQ_M and NMR under different levels of occlusion

In the second experiment, we also use Subsets 1 and 2 for training and Subset 3 for test, but with different kinds of occlusions: cup, dollar, cartoon mask, book, flower and puzzle in test images (as shown in Fig. 10). The recognition rate of each method is shown in Fig. 11. The proposed NMR achieves the best result among all methods. This experiment demonstrates that NMR is more robust than others for face recognition with different, contiguous occlusions.

In the third experiment, for the test image in Subset 3, we impose another two special occlusions: a square black block and a square random block whose elements are random numbers between 0 and

255. Fig. 12 shows the recognition rates of each method under various occlusion levels with black block and random block. In general, the results in Fig. 12 are consistent with those in Fig. 9. NMR always achieves robust performance and outperforms state-of-the-art methods in both occlusion cases. In Fig. 12 (a), the performance difference between NMR and RSC (or SSEC) is not as remarkable as that shown in Fig. 9 when the occlusion level is over 50%. The recognition rate of NMR is 57.3%, 6.2%, 4.0% higher than SRC, RSC and SSEC when the occlusion level is 60%. In Fig. 12 (b), the performance difference between NMR and RSC becomes more remarkable when the occlusion level is equal to or larger than 30%. The proposed NMR still achieves a recognition rate of 86.4% when the occlusion level is 60%, which is 4.1%, 22.8% higher than SSEC and RSC.

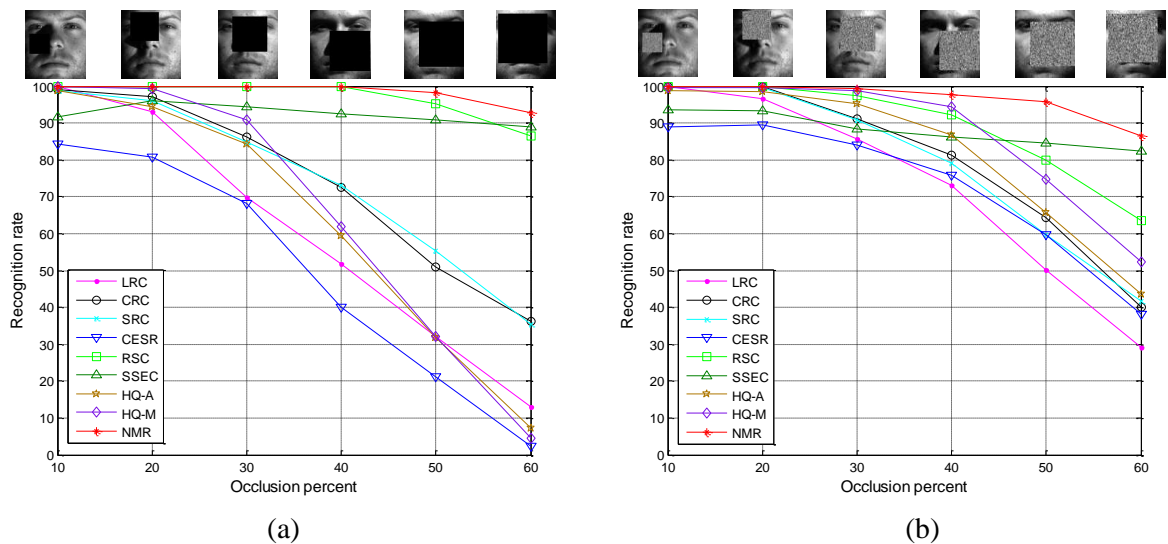

(a)                                      (b)

Fig. 12 Recognition rates (%) of LRC, CRC, SRC, CESR, RSC, SSEC, HQ_A, HQ_M and NMR under the different occlusion levels. (a) the case that test images are with the occlusion of black block; (b) the case that test images are with the occlusion of random block.

**4.3 Recognition with Real Face Disguise**

In this experiment, we evaluate the robustness of NMR in dealing with real disguise on the AR database. Here, we select 8 frontal face images without occlusion, i.e. the first 4 images of Sessions 1 and 2 for training. We construct two test sets: (i) 6 images with sunglasses from both sessions, and (ii) 6 images with scarves from both sessions. The classification results of LRC, CRC, SRC, CESR, RSC, SSEC, HQ_A, HQ_M and NMR are listed in Table 1. From Table 1, we observe that NMR achieves the highest recognition rate for each test set. For test images with sunglasses, where the occlusion level is relatively low, the sparseness assumption holds so SRC can achieve good results. Besides, in this case, CESR and HQ_M both achieve encouraging results. There is no significant performance difference between NMR and these methods. However, when the occlusion level becomes larger, in the case of images with scarves, the advantage of NMR becomes evident, i.e. NMR significantly outperforms others. It is strange that SSEC does not perform well in this experiment. The possible reason is that the test images with sunglasses and scarves were taken under different lighting conditions, e.g. left or right light on. SSEC seems to be very sensitive to illumination changes. This conjecture will be further verified in the following experiments.

Table 1 Recognition rates (%) of LRC, CRC, SRC, CESR, RSC, SSEC, HQ_A, HQ_M and NMR on the AR database

|  | LRC | CRC | SRC | CESR | RSC | SSEC | HQ_A | HQ_M | NMR |
|---|---|---|---|---|---|---|---|---|---|
| Sunglasses | 92.8 | 93.5 | 94.4 | 95.0 | 89.2 | 79.0 | 94.7 | 95.0 | *96.9* |
| Scarf | 30.7 | 63.6 | 57.6 | 33.5 | 66.8 | 49.1 | 48.7 | 50.1 | *73.5* |

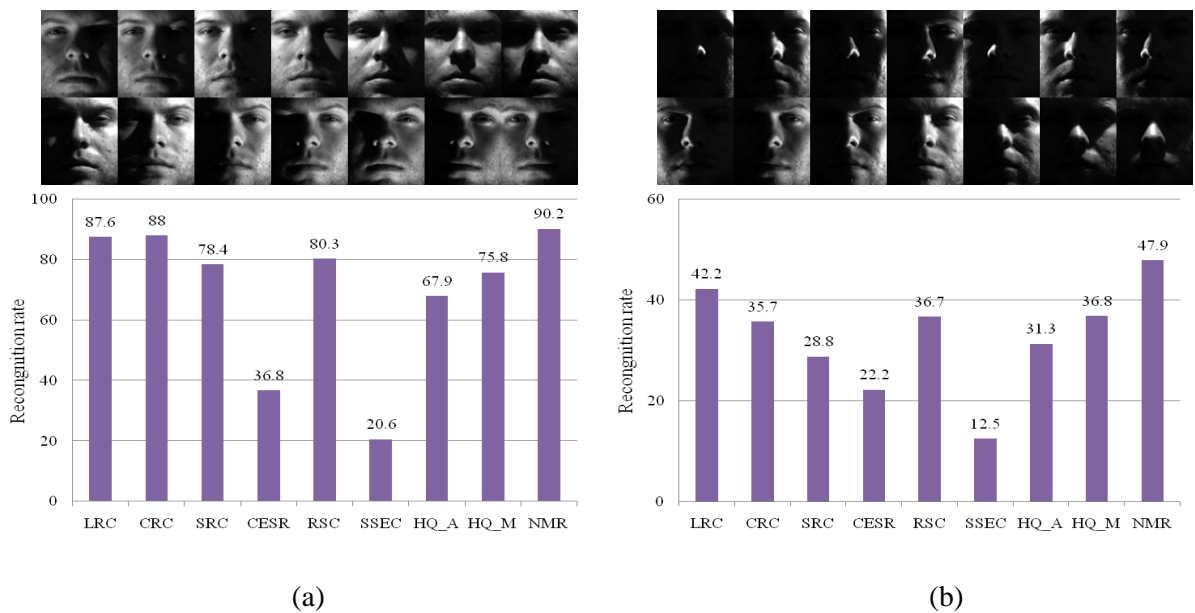

(a)                      (b)

Fig. 13 Recognition rates (%) of each classifier under different illumination conditions on the Extended Yale B database. (a) on the Subset 4, (b) on the Subset 5.

### 4.4 Recognition with different Illumination

In this section, we test the proposed method under different illumination conditions. In the first experiment we choose Subset 1 of the Extended Yale B database for training. As we know, extreme illumination change is a challenging task for most face recognition methods. Therefore, Subsets 4 and 5 with extreme lighting conditions are used for test, respectively. Fig. 13 shows the recognition rates of all methods tested on Subset 4 and Subset 5. For both subsets, the proposed NMR achieves the best results among all methods. Some robust sparse representation methods like CESR, HQ_A, HQ_M seems not very robust to extreme illumination changes. SSEC, as a method designed exclusively for contiguous occlusion, is not suitable for extreme illumination changes either. However, the classical linear regression based method, LRC, seems less sensitive to illumination changes than robust sparse representation methods.

We conducted the second experiment on the Multi-PIE database. There are 249 subjects in Session 1, and 166, 160, 175 subjects in Sessions 2, 3 and 4, respectively. All subjects of Session 1, each having 8 frontal neutral images with slight illumination changes are used for training. All subjects of Sessions 2, 3 and 4, each having 10 frontal neutral images with different illumination variations (as shown in Fig. 8) are used for test. Table 2 lists recognition rates of all methods for the three test sets. The proposed NMR always achieves the best results, but the (robust) sparse representation methods like SRC, RSC and HQ_M also achieve competitive results in these tests. Note that the illumination conditions of images in the Multi-PIE database are much better than those in the Extended Yale B database as used in the foregoing experiment. It seems that SRC, RSC and HQ_M are insensitive to relatively slight illumination changes.

Table 2 Recognition rates (%) of LRC, CRC, SRC, CESR, RSC, SSEC, HQ_A, HQ_M and NMR on the Multi-PIE database under different illuminations

|  | Session 2 | Session 3 | Session 4 |
| --- | --- | --- | --- |
| LRC | 76.4 | 67.0 | 74.2 |
| CRC | 82.4 | 71.8 | 80.2 |
| SRC | 82.7 | 73.6 | 82.0 |
| CESR | 76.6 | 64.9 | 76.2 |
| SSEC | 66.2 | 53.6 | 59.1 |
| RSC | 82.8 | 75.3 | 81.8 |
| HQ_A | 79.5 | 68.6 | 77.7 |
| HQ_M | 82.7 | 74.2 | 83.2 |
| NMR | 85.8 | 77.9 | 84.5 |

**4.5 Experiment on the FRGC database**

In this section, we choose a subset of the FRGC database, which contains 220 persons and each person has 20 images. These images are taken in different conditions such as large illumination variations, low resolution of the face region and possible blurring. We use the first 10 images per class for training, and the remaining for test. Here, the face region of each image is first cropped from the original high-resolution still images and resized to a spatial resolution of $32 \times 32$. The classification results of LRC, SRC, CRC, CESR, RSC, SSEC, HQ_A, HQ_M and NMR are shown in Table 3. These results demonstrate the effectiveness of the proposed method for face recognition in the different conditions. SSEC is designed exclusively for face recognition with contiguous occlusion, but its performance is not good in general cases without occlusion. In contrast, some other methods like CRC, RSC and HQ_M achieves very good results in this experiment.

Table 3 Recognition rates (%) of LRC, CRC, SRC, CESR, RSC, SSEC, HQ_A, HQ_M and NMR on the FRGC database

|  | LRC | CRC | SRC | CESR | RSC | SSEC | HQ_A | HQ_M | NMR |
|---|---|---|---|---|---|---|---|---|---|
| Rates | 77.0 | 92.2 | 89.2 | 81.9 | 92.0 | 70.5 | 84.7 | 91.9 | *93.3* |

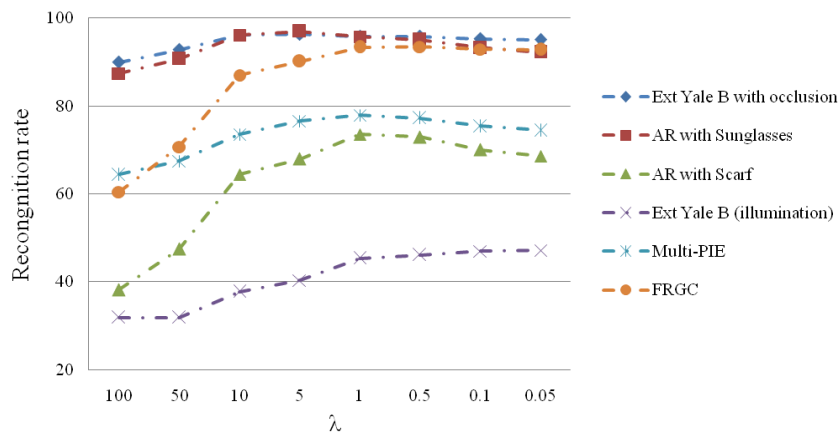

Fig. 14 Recognition rates of NMR with different regression parameters on the Extended Yale B, AR, Multi-PIE and FRGC databases

**4.6 Parameter Discussion**

In this section, we discuss how the regression parameter $\lambda$ affects the performance of our NMR method in different face recognition scenarios. For experiments with occlusions, on the Extended Yale B database, Subset 1 and 2 are used for training and Subset 3 with block occlusion (FeiFei, 50% occlusion level) for test. On the AR database, the first 4 images of each person in Session 1 and 2 are used for training and the images with sunglasses (scarf) of each session for testing. For experiments with illumination changes, the images from Subset 1 of the Extended Yale B database are used for training, and the images with extreme illuminations in Subset 5 for testing. For the Multi-PIE database, Session 1 is used for training and Session 3 for test. Additionally, we perform experiment on the FRGC database using the same experiment setting as in Section 4.4.

The recognition rates of NMR with respect to different regression parameters on four face image databases are shown in Fig. 14. From Fig. 14, we can see that NMR always achieve its optimal or nearly optimal performance in the range of [0.5 1] under different face recognition scenarios. Thus, it's easy to set the parameter of the proposed NMR in real-world applications.

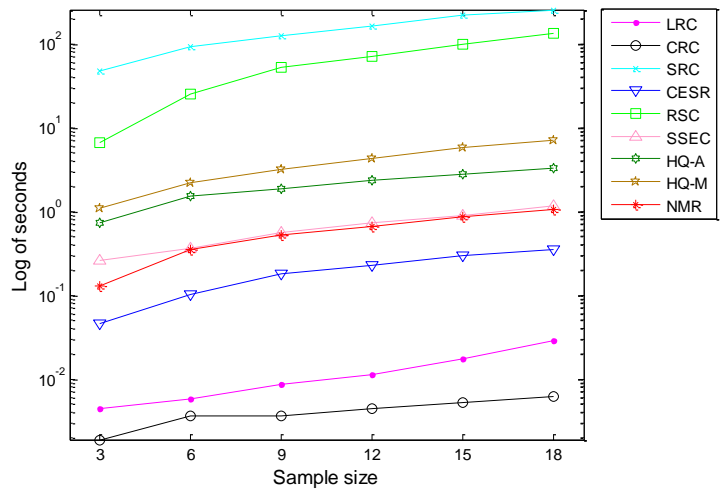

Fig. 15 Illustration of the average running time (second, in logs) of recognizing one testing sample for each method on the Extended Yale B database

## 4.7 Comparison Analysis of Running Time

In this subsection, we compare the running time of the proposed NMR with state-of-the-art methods. Our programming environment is Matlab 2011, and all algorithms are implemented on a Core Duo 2.93GHz with 4G RAM desktop. We conduct the experiment in face recognition with block occlusion (FeiFei, 50% occlusion level) on the Extended Yale B database. The number of training samples of each class varies from 3 to 18, with an interval of 3. The average running time (second, in logs) of recognizing one testing sample for each method are illustrated in Fig. 15.

From Fig. 15, we can see that LRC and CRC are the fastest methods, because they only involve a linear regression problem which has a close-form solution. But, the two methods are not very robust, particularly when there are extreme occlusion levels. CESR is also faster than NMR, but its recognition performance is always remarkably lower than NMR. SSEC performs as fast as NMR, but SSEC is designed exclusively for contiguous occlusion; it is sensitive to illumination changes. In contrast, the proposed NMR is a more general face recognition algorithm. Compared to the halt-quadratic based sparse representation methods HQ_A and HQ_M, NMR is faster and more robust to occlusion and illumination changes. The other robust methods such as RSC and SRC, are significantly more time-consuming than NMR. The empirical computational complexity of RSC is $O(t(n^2 m^{1.3}))$, where $t$ is the iteration number of RSC, while that of SRC is $O(n^2(m+n)^{1.3})$ because it needs to use an extra identity matrix to represent the occluded or corrupted pixels [7, 4]. NMR has a computational complexity of $O(k(m^{1.5}+mn))$, where $k$ is the number of iterations. In general, the computational complexity of NMR is much lower than those of RSC and SRC.

## 5. Conclusions and Future Work

This paper presents a nuclear norm based matrix regression (NMR) model and uses the augmented Lagrange multipliers method to calculate the regression coefficients. The convergence analysis of the proposed algorithm is given. The proposed NMR classifier is examined on four popular face image databases: the Extended Yale B, AR, Multi-PIE and FRGC, and experimental results indicate that (i) NMR is more robust than state-of-the-art regression based methods for face recognition with occlusions and illumination changes, and (ii) NMR is more powerful than the structured sparse error coding model, which was designed exclusively for dealing with occlusion but turns out to be very sensitive to illumination changes, (iii) NMR can achieve satisfying results for general face recognition tasks without occlusion.

Although the NMR is faster than many robust regression methods, its speed is still a big constraint for real-world applications. How to design a computationally more efficient algorithm for solving the nuclear norm based matrix regression is an open problem. In addition, our model uses the nuclear norm as a criterion to capture the low-rank structural noise. Whether this model is effective for more complex noise or how to extend the model for general noise needs further investigation.

**Acknowledgments**

This work was partially supported by the National Science Fund for Distinguished Young Scholars under Grant No. 61125305, and the Key Project of Chinese Ministry of Education under Grant No. 313030.

**Jian Yang** received the BS degree in mathematics from the Xuzhou Normal University in 1995. He received the MS degree in applied mathematics from the Changsha Railway University in 1998 and the PhD degree from the Nanjing University of Science and Technology (NUST), on the subject of pattern recognition and intelligence systems in 2002. In 2003, he was a postdoctoral researcher at the University of Zaragoza. From 2004 to 2006, he was a Postdoctoral Fellow at Biometrics Centre of Hong Kong Polytechnic University. From 2006 to 2007, he was a Postdoctoral Fellow at Department of Computer Science of New Jersey Institute of Technology. Now, he is a professor in the School of Computer Science and Technology of NUST. He is the author of more than 80 scientific papers in pattern recognition and computer vision. His journal papers have been cited more than 2000 times in the ISI Web of Science, and 4000 times in the Web of Scholar Google. His "2DPCA" paper published in TPAMI 2004 has been cited more than 2000 in Scholar Google. His research interests include pattern recognition, computer vision and machine learning. Currently, he is an associate editor of Pattern Recognition Letters and IEEE Trans. Neural Networks and Learning Systems, respectively.